# Layer Pruning for Accelerating Very Deep Neural Networks

Weiwei Zhang, Changsheng chen, Xuechun Wu, Jialin Gao, Di Bao, Jiwei Li, Xi Zhou

Cloudwalk Technology, Shanghai, China.

{zhangweiwei,chenchangshen,wuxuechun,gaojialin,baodi,lijiwei,zhouxi}@cloudwalk.cn

# 1 Abstract

In this paper, we propose an adaptive pruning method. This method can cut off the channel and layer adaptively. The proportion of the layer and the channel to be cut is learned adaptively. The pruning method proposed in this paper can reduce half of the parameters, and the accuracy will not decrease or even be higher than baseline.

# 2 Introduction

With the deepening of deep learning network, more and more achievements have been made in computer vision and other fields, and the accuracy is higher and higher. At present, many front-end devices, such as robots, cameras, UAVs, apps and a series of resources are insufficient, but want to share the deep learning results urgently need an algorithm that can greatly reduce the amount of computation and maintain the original accuracy. Pruning is one of the effective algorithms, which can reduce the volume of the model, reduce the amount of model parameters, and improve the speed of inference. Relevant researchers have done a lot of work in this field. We propose an adaptive layer pruning method of multi task joint training, which can cut off some layers of neural network adaptively, improve inference speed, reduce model volume and parameters, at the same time, it can ensure that the accuracy basically retains the original accuracy or even higher than the original accuracy. At the same time, it is also a way to replace the original multi-layer convolution kernel with 1x1 convolution kernel when the information is thin.

## 3 Related work

With the development of deep learning algorithm, more and more deep learning models are used in various scenarios, including some front-end devices. But front-end devices usually don't have enough resources like servers. When running a large deep learning model, we usually need to make some improvements to adapt to the front-end devices. The goal of improvement can be roughly divided into three points: the first point is to improve the speed of model inference. The second point is to reduce the memory or explicit memory consumed by the model in the inference process [6]. The third is to reduce the volume of the model. At present, there are many ways to solve these problems mentioned above. The first direction is low precision model training, such as the method of quantifying network weight parameters represented by binary weights or int8[3]. This method can solve the three problems mentioned above. However, since the network weight is quantified as binary, special bit wise acceleration library or special hardware is needed for inference, so it is not convenient for deployment, and a lot of customization related work needs to be done. The second direction is quantification, which focuses more on using some clustering centers to represent the weight parameters. Common clustering algorithms include PQ, VQ, kmeans and so on. This method can solve the last problem mentioned above. However, when inference, because multiple parameters are represented by the same center point, we need to restore the center point in the way of one to many mapping. In this way, in the process of inference, it can not reduce the run-time-memory and it is difficult to improve the inference speed, only reducing the size of the model volume [3]. In the actual application scenario, I think the most important problem is the inference speed of the model, so this method is not good enough. The fourth direction is to design a new model structure. The typical work is to use squeezenet [7], nasnet [8], shufflenet [9,10], and mobilenet [11,12]. This kind of network structure is designed to be very small but relatively high precision. This is mainly due to the application of a large number of separable convolutions. This method is generally better, but it needs too much human design and participation. The fourth method is pruning. The first way to prune is to sparse the weight of a channel in a certain layer, then turn the channel into a sparse matrix, and then use some sparse matrix acceleration libraries to accelerate, but this method is too cumbersome and not easy to deploy. At present, channel pruning is the main method, that is to cut off all the unimportant channels, so that the three problems mentioned above can be solved without other tedious operations, and it is convenient to deploy without relying on special hardware and acceleration library. We propose an adaptive pruning method, which can prune both channels and layers. The proportion of channels and layers to be pruned is learned by the model itself in training. We use the learning method of multi task mutual promotion to combine the first two steps of the original three-step iterative training.

## 4 Approach

## 4.1 BN layers

In order to make the model more effective, BN layer is usually added to the neural network. The function of BN layer can be expressed by the

following formula:

$$y = \alpha \times \frac{x - u}{\sqrt{6^2 + \varepsilon}} + \beta$$

Where $\alpha$ and $\beta$ are scaling coefficients, $u$ and $6^2$ are statistical mean and variance.

We add L1 weight penalty to $\alpha$, forcing the network to make the $\alpha$ value of some channels tend to 0 in the process of learning. We think that the channels corresponding to $\alpha$ tend to 0 are relatively unimportant. We cut off the front and back channels corresponding to those $\alpha$ which tend to 0, so as to achieve the purpose of pruning.

## 4.2 LOSS

Our loss function is based on the normal training loss function and adds a penalty to α, which is expressed as follows:

$$Loss_{total} = Loss_{normal} + \lambda \times f(\alpha)$$

Where $f(\alpha)$ stands for the norm penalty to $\alpha$, where L1 norm is taken, and $\lambda$ stands for the proportion weight of the penalty to $\alpha$.

## 4.3 Multi task learning

We use the learning method of multi task mutual promotion to combine the first two steps of the original three-step iterative training into one. The specific training steps are changed as follows:

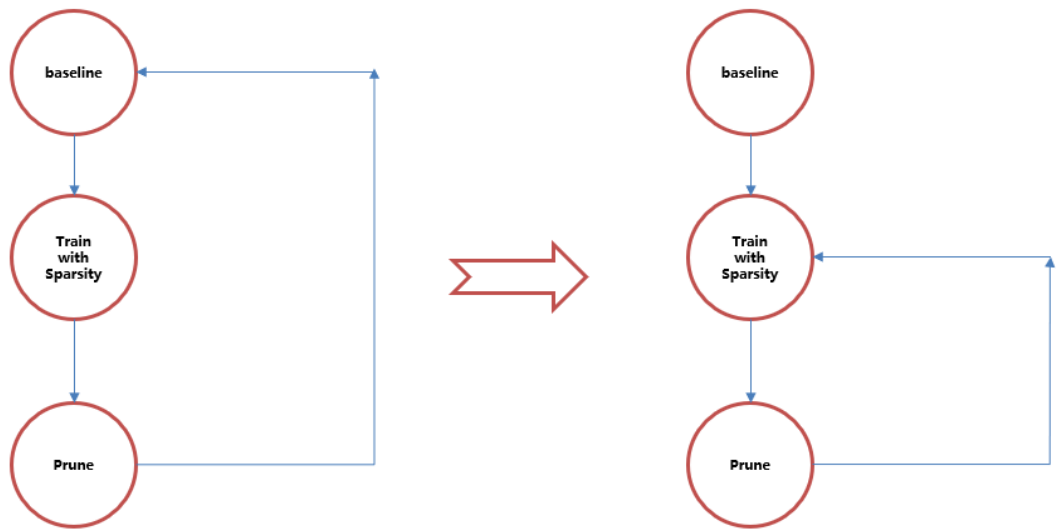

For the model after channel or layer pruning, we also conduct sparsity operation at the same time of finetune to prepare for the next pruning. In other words, we can do fine tune and sparse at the same time, and the two tasks promote each other.

## 4.4 1x1 convolution kernel as a bridge for channel merging

We set a certain pruning threshold and continue to cut the channel iteratively. Adjust the threshold value so that there is only one channel left for the α of some BN layers. Then we connect the convolution layer conv before and after the BN layer with 1x1 convolution kernel.

For example, the shape of conv1 on the previous layer is:

$$in\_channel1 \times out\_channel1 \times kernel\_size1 \times kernel\_size1$$

The shape of BN layer is:

$$in\_channel2$$

The shape of the latter convolution conv2 is:

$$in\_channel2 \times out\_channel2 \times kernel\_size2 \times kernel\_size2$$

We combine the original conv1, BN and cnov2 into a new convolution layer, named new_conv, with the shape as follows:

$$in\_channel1 \times out\_channel2 \times 1 \times 1$$

In other words, the convolution kernel of 1x1 is used to replace the original three layers, which not only plays the role of connecting channels, but also reduces many parameters.

### 4.4.1 init weight

$$conv_{new}.weight = 0$$

In other words, the weight of all channels in the new convolution kernel becomes 0, which is equivalent to a residual of 0. In this way, we can only keep the bypass trunk branch and remove the whole residual block.

### 4.4.2 The second method of initializing weights

The weight value of this 1x1 convolution kernel initialization, namely $conv_{new}.weight$, can be calculated with the following pseudo code:

$$weight = conv1.weight$$
$$weight = BN(weight)$$
$$weight = ReLU(weight)$$
$$weight = conv2(weight)$$
$$conv_{new}.weight = weight$$

In this way, some layers in the residual block are removed and a new 1x1 convolution kernel is used instead of connecting the front and back layers.

### 4.4.3 Understanding of initialization weight method

The method in 4.4.1 can be understood as that the information of a certain layer in the residual block is very thin, closing to 0, so we can ignore the influence of the whole residual block on the network, only keep the bypass branch, which can greatly improve the inference speed of the network without losing the accuracy.

In 4.4.2, $conv1$, BN, and $conv2$ are all linear layers, while relu can be regarded as a nonlinear layer composed of two linear layers. If a, b and c are all linear, then the combination law is satisfied.

$$abc = a(bc)$$

if:

$$out1 = (input.mm.conv1).mm.BN.mm.relu.mm.conv2$$

assume:

$$conv_{new} = conv1.mm.BN.mm.relu.mm.conv2$$

then:

$$out2 = input.mm.conv_{new}$$

If relu is not taken into account, then other layers are linear, then out1 and out2 calculated above are equivalent. So we can use the new 1x1 convolution kernel instead of the original 2-layer convolution kernel. We can remove the original only one channel, that is, the convolution kernel with sparse information and connect it in a new way. In this way, the parameters are reduced, the reasoning speed is increased, and the accuracy is not affected.

But we note that relu is nonlinear:

$$f(x) = \begin{cases} x & if\ x > 0 \\ \lambda x & if\ x \leq 0 \end{cases}$$

If all $x > 0$, then x will keep the original value of normal propagation, at this time, the above linear condition is met. This is equivalent to removing some layers from the residual block while retaining some layers.

If all $x \leq 0$, the propagation value of the residual part is 0, then the initialization weight will be reduced to the case that the whole residual block is discarded in 4.4.1, which is equivalent to that only the bypass branch has information propagation.

If part $x > 0$, Part $x \leq 0$, then matrix multiplication can be regarded as adding some perturbations in the case of $x > 0$ linear propagation. At this time, the calculated out2 value is no longer equivalent to the original out1. In matrix multiplication, some values of the left matrix become 0, and the calculated out2 can be regarded as some perturbations based on out1. Because the information here is relatively thin and unimportant, some disturbances do not affect the result of inference. In this way, the effect of disturbance is better than that of random initialization of $conv_{new}$.

In summary, the role of relu is to choose whether to remove the whole residual block or retain part of the layer. Removing the whole residual block shows that the information of the residual block is not very valuable. Some of the layers in the residual block are reserved, which is equivalent to the weight initialization of $conv_{new}$, satisfying the matrix combination law or making some perturbations in the case of linear combination law. With the existence of $relu$, the weight of $conv_{new}$ has three states: satisfying the combination law, removing the whole residual block, and making some disturbances based on the combination law. Either way, it may be what the network itself wants. The weight initialization methods of 4.4.1 and 4.4.2 are all available. 4.4.1 is relatively simple in operation, and the removal of the whole residual block can also improve the inference speed. 4.4.2 that is to say, removing the layer in some residual blocks can retain some original micro information, which can improve the effect to some extent, but the acceleration is not as obvious as the effect in 4.4.1. No matter which initialization method, the final iteration operation of finevalue and sparsity is required.

This method can be applied to any network with BN layer and can also be extended to the network without BN layer. We only need to use 1 x 1 convolution kernel to combine the front and back layers of only one channel. This method has nothing to do with specific tasks. It can be applied to recognition tasks, detection tasks, segmentation and other tasks.

## 5 Experiment

### 5.1 Datasets

We use cifar-10 [4] as our dataset to verify our approach. Cifar dataset is a natural image with a resolution of 32 * 32. Cifar-10 is a data set composed of 10 categories extracted from cifar data set, including 50000 training samples and 10000 test samples. Some conventional image enhancement strategies are adopted by us.

### 5.2 Models

We use the RESNET network with a depth of 164 for training and testing [5]. Bottleneck structure is used in this network. When the layer is cut off, we will change the structure of bottleneck.

### 5.3 Evaluation metrics

We use the following metrics to measure the model: (1) Top1 (2) parameters (3) model size (4) inference speed.

### 5.4 Results and Discussions

Experimental results of similar work[2]:

| Test Errors on CIFAR-10[2] | | | | |
|---|---|---|---|---|
| Model | Top1 (%) | Parameters | Model size | Inference ratio |
| ResNet-164(Baseline) | 94.58 | 1.70M | 13.3M | 1x |
| ResNet-164(40% Pruned) | 94.92 | 1.44M | | |
| ResNet-164(60% Pruned) | 94.73 | 1.10M | 4.87M | 1.5x |

Our experimental results:

| Test Errors on CIFAR-10 | | | | |
|---|---|---|---|---|
| Model | Top1 (%) | Parameters | Model size | Inference ratio |
| ResNet-164(Baseline) | 94.58 | 1.70M | 13.3M | 1x |
| ResNet-164(40% Pruned) | 95.09 | 1.44M | | |
| ResNet-164(60% Pruned) | 95.18 | 1.10M | 4.87M | |
| ResNet-164(40% Pruned) | 95.02 | 0.95M | 3.5M | 1.7x |

First of all, we can see from the table that our pruning rate is larger than that in [2], but our Top1 and inference speed are better than that in [2], and our Top1 is even better than that in baseline. This shows that our iterative layer pruning training removes the redundant parameters of the model, but improves the generalization ability of the model, which shows that our layer pruning strategy is effective.

In addition, we noticed a phenomenon that the pruned model of [2] appeared the inflection point too early. Generally speaking, it is a normal phenomenon that the accuracy of the model declines after pruning. We can also improve the accuracy through finetune, but it is a bad phenomenon that the inflection point appears too early, which shows that the model is not strong enough. This phenomenon also shows that the shearing strategy we proposed is more reasonable from another point of view.

Thirdly, our parameter size, inference speed and model size are better than those of our peers.

In literature [1], a channel pruning method is proposed. Although it has achieved good results, the author embeds three spp modules between the 5th and 6th convolution layers of the backbone network of yolov3, so it is not clear why the effect of this paper is improved. Our algorithm does not use any external extra modules to improve our effect. Our adaptive pruning can remove redundant channels and layers of the network, and improve the generalization ability of the network adaptively. In addition to the strategies mentioned in our paper, no additional operations are added. It can be seen that our results are better than those of our peers.

# 6 Conclusion

We propose an adaptive pruning method, which can be applied to any network with BN layer and any network without BN layer. This strategy has nothing to do with specific tasks, and is suitable for a series of tasks such as recognition, detection and segmentation.